\documentclass[conference]{IEEEtran}
\IEEEoverridecommandlockouts
\usepackage{cite}
\usepackage{amsmath,amssymb,amsfonts}
\usepackage{algorithmic}
\usepackage{graphicx}
\usepackage{textcomp}
\usepackage{xcolor}
\def\BibTeX{{\rm B\kern-.05em{\sc i\kern-.025em b}\kern-.08em
    T\kern-.1667em\lower.7ex\hbox{E}\kern-.125emX}}

\usepackage{adjustbox}
\usepackage{array}
\usepackage{amsthm}
\usepackage{booktabs}
\usepackage{cleveref}
\usepackage[inline]{enumitem}
\usepackage[utf8]{inputenc}
\usepackage{mathtools}
\usepackage{multirow}
\usepackage{pgfplots}
\usepackage[eulergreek]{sansmath}
\usepackage{siunitx}
\usepackage{url}
\usepackage{tabularx}
\usepackage{tikz}
\usepackage{tkz-euclide}
\usepackage{xspace}
\definecolor{hgf-blue}{RGB}{0,90,160}
\definecolor{hgf-green}{RGB}{140,180,35}
\definecolor{hgf-gray}{RGB}{90,105,110}
\definecolor{hgf-purple}{RGB}{160,35,90}
\definecolor{hgf-red}{RGB}{210,50,100}
\definecolor{hgf-orange}{RGB}{240,120,30}
\definecolor{hgf-yellow}{RGB}{255,210,40}
\definecolor{hgf-turqoise}{RGB}{80,200,170}
\definecolor{hgf-dark-blue}{RGB}{50,100,105}

\pgfplotsset{
    compat=newest,
    discard if not/.style 2 args={
        x filter/.code={
          \edef\tempa{\thisrow{#1}}
            \edef\tempb{#2}
          \ifx\tempa\tempb
          \else
            
          \fi
        }
    }
}

\usetikzlibrary{
    arrows, 
    calc,
    intersections,
    pgfplots.groupplots,
    positioning,
    shapes.geometric,
}

\newcommand{\fancyname}{ReCycle}
\def\decoding{SPD}

\def\datalength{L}
\def\daylength{D}
\def\metafeatures{M}
\def\daycats{K}
\def\pslpwindow{k}
\def\forecastwindow{F}
\def\historicwindow{H}
\def\batchsize{B}
\def\genericLength{N}

\def\featuredimension{f}

\def\entsoe{ENTSO-E}
\def\entsoede{ENTSO-E}
\def\water{Water}
\def\etth{ETT}

\def\etthtwo{ETTh2}
\def\ucipt{ELD}
\def\traffic{Traffic}

\def\historicprofile{HP}
\def\localprofile{RHP}

\begin{document}

\title{ReCycle: Fast and Efficient Long Time Series Forecasting with Residual~Cyclic~Transformers
\thanks{
This work is supported by the Helmholtz Association Initiative and Networking Fund under the Helmholtz AI platform grant and the HAICORE@KIT partition and by the German Federal Ministry of Education and Research under the 01IS22068 - EQUIPE grant. The authors gratefully acknowledge the computing time made available to them on the high-performance computer HoreKa at the NHR Center KIT. This center is jointly supported by the Federal Ministry of Education and Research and the state governments participating in the NHR (www.nhr-verein.de/unsere-partner).
\\\newline
\textcopyright 2024 IEEE. Personal use of this material is permitted. Permission
from IEEE must be obtained for all other uses, in any current or future
media, including reprinting/republishing this material for advertising or
promotional purposes, creating new collective works, for resale or
redistribution to servers or lists, or reuse of any copyrighted
component of this work in other works.
}}

\author{\IEEEauthorblockN{Arvid Weyrauch}
\IEEEauthorblockA{%\textit{Scientific Computing Center} \\
\textit{Karlsruhe Institute of Technology (KIT)}\\
%Eggenstein-Leopoldshafen, Germany \\
Karlsruhe, Gemany\\
arvid.weyrauch@kit.edu}
\and
\IEEEauthorblockN{Thomas Steens}
\IEEEauthorblockA{%\textit{Institute of Networked Energy Systems} \\
\textit{German Aerospace Center (DLR)}\\
Oldenburg, Germany \\
 t.steens@hotmail.com}
\and
\IEEEauthorblockN{Oskar Taubert}
\IEEEauthorblockA{%\textit{Scientific Computing Center} \\
\textit{Karlsruhe Institute of Technology (KIT)}\\
Karlsruhe, Gemany\\
oskar.taubert@kit.edu}
\and
\IEEEauthorblockN{Benedikt Hanke}
\IEEEauthorblockA{%\textit{Institute of Networked Energy Systems} \\
\textit{German Aerospace Center (DLR)}\\
Oldenburg, Germany \\
benedikt.hanke@dlr.de}
\and
\IEEEauthorblockN{Aslan Eqbal}
\IEEEauthorblockA{\textit{INENSUS GmbH} \\
Goslar, Germany \\
a.eqbal@inensus.com}
\and
\IEEEauthorblockN{Ewa Götz}
\IEEEauthorblockA{%\textit{Digital Industries} \\
\textit{Siemens AG, Digital Industries\hspace{2em}}\\
Karlsruhe, Germany \\
ewa.goetz@siemens.com}
\and
\IEEEauthorblockN{Achim Streit}
\IEEEauthorblockA{%\textit{Scientific Computing Center} \\
\textit{Karlsruhe Institute of Technology (KIT)}\\
Karlsruhe, Gemany\\
achim.streit@kit.edu}
\and
\IEEEauthorblockN{Markus Götz}
\IEEEauthorblockA{%\textit{Scientific Computing Center} \\
\textit{Karlsruhe Institute of Technology (KIT)}\\
Karlsruhe, Gemany\\
markus.goetz@kit.edu}
\and
\IEEEauthorblockN{Charlotte Debus}
\IEEEauthorblockA{%\textit{Scientific Computing Center} \\
\textit{Karlsruhe Institute of Technology (KIT)}\\
Karlsruhe, Gemany\\
charlotte.debus@kit.edu}
}

\maketitle

\begin{abstract}
Transformers have recently gained prominence in long time series forecasting by elevating accuracies in a variety of use cases.
Regrettably, in the race for better predictive performance the overhead of model architectures has grown onerous, leading to models with computational demand infeasible for most practical applications.
To bridge the gap between high method complexity and realistic computational resources, we introduce the Residual Cyclic Transformer, \fancyname{}. 
\fancyname{} utilizes primary cycle compression to address the computational complexity of the attention mechanism in long time series. 
By learning residuals from refined smoothing average techniques, \fancyname{} 
surpasses state-of-the-art accuracy in a variety of application use cases.
The reliable and explainable fallback behavior ensured by simple, yet robust, smoothing average techniques additionally lowers the barrier for user acceptance.
At the same time, our approach reduces the run time and energy consumption by more than an order of magnitude, making both training and inference feasible on low-performance, low-power and edge computing devices.
Code is available at \url{https://github.com/Helmholtz-AI-Energy/ReCycle}
\end{abstract}


\section{Introduction}

Among the different applications of machine learning (ML) methods, time series forecasting is one of the most widely encountered and most complex tasks.
%The ability to predict temporal behavior is the key to technological innovation. 
%Due to the threats of climate change and the related factors in our socio-economic life, forecasting temporal behavior within many domains, and specifically those interested in resource management, remains a pressing yet unresolved challenge.
While for a long time, traditional statistical methods, such as smoothing averages or auto-regressive moving averages, dominated the algorithmic landscape for analysis and prediction of temporal behavior, recent advances in deep learning (DL), and natural language processing (NLP) in particular, have paved the way for neural network-based approaches~\cite{lara2021experimental}. 
%The superior predictive performance of neural networks allows researchers to address some of the most eminent tasks in a variety of fields, such as computer vision and natural language processing. 
The introduction of the Transformer architecture~\cite{vaswani_attention_2017} has preluded significant breakthroughs in sequence processing, making it a natural candidate for state-of-the-art (SOTA) time series forecasting. 
However, adaptation to time series applications poses a mayor challenge: The calculation and memory complexity of the attention mechanism for a sequence of length $L$ is $\mathcal{O}(L^2)$, making the forecasting of long time series exceptionally compute intensive. For real-time deployment on hardware with limited memory and computation capabilities, this computational footprint is prohibitive. Moreover, dot-product attention was originally designed for multi-feature tokens, complicating the method transfer to uni-variate time-series.

Several authors have put forward approaches for general Transformer-based time series forecasting problems that attempt to tackle the computational complexity of the attention mechanism~\cite{zhou_informer_2021,wu_autoformer_nodate,zhou_fedformer_2022}. 
However, while they theoretically reduce the complexity to logarithmic or even linear, practical implementations do not yield significant computational performance increase, as shown by our experiments. The increased computational overheads and complex model structures ultimately yield longer run times and consequently, higher energy consumption. This renders them infeasible for typical deployment hardware, such as on-board systems, embedded devices, or low-power hardware in sensors. But more importantly, it contributes to the steadily increasing environmental footprint of AI methods that has accompanied the drive for more accurate models, a trend which has been termed Red AI~\cite{schwartz2020green}.
%For computer vision networks, this problem has been addressed~\cite{howard2017mobilenets, dong2022tinynet}, but it remains unanswered for Transformer architectures in time series forecasting.

In this work, we present \fancyname{}: Residual Cyclic Transformers, a method for fast and energy efficient time series forecasting. Our contributions include a rigorous mathematical discussion of the attention mechanism for single-feature sequences, leading to something that we term \textit{scalar breakdown of dot-product attention}.
The corresponding incapability of the attention mechanism to capture relational properties between single-featured tokens has direct implications for univariate time series forecasting, which adds to our motivation for this work.

We propose two conceptual changes in representing temporally resolved data, to tackle the growing computational demand in long-time series forecasting with Transformer-based architectures:
\begin{enumerate}
    \item Primary Cycle Compression (PCC): Real-world time series applications often exhibit distinct patterns and characteristics, such as daily, weekly, and seasonal cycles. We use this fact to our advantage, converting univariate time series into multivariate time series over these cycles. PCC addresses the scalar breakdown of dot-product attention in an easy and intuitive way. Furthermore, it naturally bypasses the memory and computation bottleneck, leading to simpler, yet computationally faster and more energy efficient time series forecasting.
    \item \textit{Residual Learning}: We leverage prior knowledge on periodicity and underlying temporal profiles, and accounts for temporal locality such as concept drift through \textit{recent historic profiles}, a new form of smoothing average naive forecast. By learning only differences to these RHP, i.e., residuals, we achieve improved prediction accuracy as the model can focus specifically on these modulations instead of the often predominant but already known periodicity.
\end{enumerate}

We evaluate our proposed method on two representatives SOTA Transformer-based architectures for a number of different datasets, demonstrating that \fancyname{} can be easily integrated with existing Transformer-based architectures, leading to superior forecasting while requiring a fraction of their run time and energy consumption. We further show, that with usage of \fancyname{}, Transformer-based architectures clearly outperform current non-Transformer models.

\section{Related Work}
Neural methods for long time series forecasting have undergone massive development since the publication of the Transformer architecture~\cite{vaswani_attention_2017}.
A comprehensive review can be found in \cite{wen2022transformers}.
LogTrans~\cite{li2019enhancing} was the first notable adaptation of the Transformer specifically for time series forecasting.
To address the issue of local context insensitivity of the self-attention mechanism, the authors substituted the point-wise dot-product with causal convolutions.
A sparse bias in form of a LogSparse mask was used to reduce computational complexity to $\mathcal{O}(L\,\log L)$.
Informer~\cite{zhou_informer_2021} focuses on dimensionality reduction via random subsampling of attention queries.
A metadata input representation related to positional encoding is used to transform univariate time-series into multivariate ones.
Autoformer~\cite{wu_autoformer_nodate} introduced a local mean-based decomposition method and replaces the dot product attention with an auto-correlation mechanism based on Fourier transforms for lower complexity.
FEDformer~\cite{zhou_fedformer_2022} combines these ideas by selecting a fixed number of Fourier modes for auto-correlation and introducing a decomposition scheme using multiple filter lengths.
Pyraformer~\cite{liu2021pyraformer} proposed the pyramidal attention module that uses inter-scale tree structures and intra-scale neighboring connections to leverage multi-resolution representations of time series. 
The recently introduced Crossformer~\cite{zhang2022crossformer} focuses on multivariate time-series, leveraging cross-dimension dependencies through dimension-segment-wise embedding and a two-stage attention layer.

Recent work on Transformer-based architectures for long time series forecasting has investigated approaches that are similar to individual components of \fancyname{}.
The ETSFormer~\cite{woo2022etsformer} exploits exponential smoothing attention and frequency attention to replace the self-attention mechanism in vanilla Transformers, thus improving both accuracy and efficiency.
PatchTST~\cite{nie2022time} proposes an approach not too different from our PCC, by segmenting the time series into subseries-level patches which are used as input tokens to Transformer. Furthermore, PatchTST promotes channel-independence for multivariate time-series, attributing the same embedding and Transformer-weights to each channel.

While these models address many of the inherent challenges for the application of Transformers to time series, their approaches are designed for generic time series and thus do not leverage any problem-specific features or properties. Moreover, their typically highly complex architectures, used to make the model generalize better, introduce significant practical implementation overhead and higher demand for computational resources. 
A recent study questions the feasibility and efficiency of Transformers for time series forecasting, indicating that linear models might yield better predictive performances~\cite{zeng_are_2022}.
In response to the excessive demand in computational resources, researchers have turned to model approaches based on multi-layer perceptrons (MLP), such as NBEATS~\cite{oreshkin2019n} or NHiTS~\cite{challu2023nhits}, showing on-par and even improved accuracy compared to Transformer-based approaches. 

\section{Notation}
\label{sec:notation}

\begin{figure*}[htb]
    \centering
    \begin{tikzpicture}
    \sffamily
    \begin{axis}[
        width=0.34\linewidth,
        height=0.25\linewidth,
        axis x line=left,
        axis y line=left,
        xlabel=time,
        ylabel=value,
        label style={font=\tiny\bfseries},
        scaled y ticks=false,
        xtick={0, 24, 48, 72, 96, 120, 144},
        xticklabels=\empty,
        yticklabels=\empty
    ]
        \addplot[thick, color=hgf-blue] table[col sep=comma, x=step, y=original] {images/pcc-data.csv};
    \end{axis}
\end{tikzpicture}\qquad\begin{tikzpicture}
    \sffamily
    \foreach \day [count=\x, evaluate={\c={60 / 7 * (9 - \x)}}] in {0, ..., 6} {
        \node[
            anchor=north west,
            draw=black, 
            fill=hgf-blue!\c!white,
            xshift=\x / 7 * 0.23\linewidth,
            inner sep=0,
            minimum height=0.2cm,
            minimum width=1 / 7 * 0.21\linewidth,
        ] (day\day) {};
    }

    \begin{scope}[shift={(day0.north west)}, yshift=0.1cm]
        \begin{axis}[
            name=raw,
            width=0.32\linewidth,
            height=0.15\linewidth,
            axis x line=left,
            axis y line=left,
            label style={font=\tiny\bfseries},
            ticks=none,
            clip=false,
            legend columns=3,
            legend style={
                draw=none,
                font=\tiny,
                at={(0.5, 1.7)},
                anchor=north
            }
        ]
            \addplot[
                thick, 
                draw=hgf-blue,
            ] table[col sep=comma, x=step, y=original] {images/pcc-data.csv};

            \addlegendentry{Original};
            \addlegendimage{color=hgf-red, thick};
            \addlegendentry{RHP};
            \addlegendimage{color=hgf-green, thick};
            \addlegendentry{Residuals};
            
            \draw[densely dashed, color=hgf-gray!80!white] (23,0 |- current axis.south) -- (23,0 |- current axis.north);
            \draw[densely dashed, color=hgf-gray!80!white] (47,0 |- current axis.south) -- (47,0 |- current axis.north);
            \draw[densely dashed, color=hgf-gray!80!white] (71,0 |- current axis.south) -- (71,0 |- current axis.north);
            \draw[densely dashed, color=hgf-gray!80!white] (95,0 |- current axis.south) -- (95,0 |- current axis.north);
            \draw[densely dashed, color=hgf-gray!80!white] (119,0 |- current axis.south) -- (119,0 |- current axis.north);
            \draw[densely dashed, color=hgf-gray!80!white] (143,0 |- current axis.south) -- (143,0 |- current axis.north);
        \end{axis}
    \end{scope}

    \foreach \day [count=\x, evaluate={\c={60 / 7 * (9 - \x)}}] in {0, ..., 6} {
        \node[
            anchor=north west,
            draw=black, 
            fill=hgf-blue!\c!white,
            xshift=2 / 7 * 0.23\linewidth,
            yshift=(\x + 1.5) * -0.24cm,
            inner sep=0,
            minimum height=0.2cm,
            minimum width=1 / 7 * 0.21\linewidth,
        ] (pcc\day) {};
    }

    \draw[
        -stealth, 
        shorten >=0.04cm,
        shorten <=0.04cm
    ] (day0.south) |- (pcc3.west) node [midway, left] {\tiny\bfseries PCC};

    \begin{scope}[shift={(pcc6.south east)}, xshift=0.1cm]
        \begin{axis}[
            axis x line=middle,
            axis y line=middle,
            axis z line=middle,
            y dir=reverse,
            width=5.5 / 7.0 * 0.32\linewidth,
            height=3.4cm,
            ymin=0,
            zmin=42000,
            xmax=25,
            xlabel=time,
            ylabel=cycle,
            zlabel=value,
            ticks=none,
            area plot/.style={
                draw=hgf-blue,thick,
                fill opacity=0.75,
            },
            view={50}{60},
            label style={
                font=\fontsize{5}{5}\selectfont\bfseries,
                yshift=0.4em
            },
        ]
            \addplot3[
                area plot,
                fill=hgf-blue!60!white,
            ] table[
                col sep=comma, 
                x=dstep, 
                y expr=0,
                z=original,
                discard if not={day}{0}
            ] {images/pcc-data.csv};
            \addplot3[
                area plot,
                fill=hgf-blue!51.42857142857143!white,
            ] table[
                col sep=comma, 
                x=dstep, 
                y expr=1,
                z=original,
                discard if not={day}{1}
            ] {images/pcc-data.csv};
            \addplot3[
                area plot,
                fill=hgf-blue!42.85714285714286!white,
            ] table[
                col sep=comma, 
                x=dstep, 
                y expr=2,
                z=original,
                discard if not={day}{2}
            ] {images/pcc-data.csv};
            \addplot3[
                area plot,
                fill=hgf-blue!34.28571428571429!white,
            ] table[
                col sep=comma, 
                x=dstep, 
                y expr=3,
                z=original,
                discard if not={day}{3}
            ] {images/pcc-data.csv};
            \addplot3[
                area plot,
                fill=hgf-blue!25.71428571428571!white,
            ] table[
                col sep=comma, 
                x=dstep, 
                y expr=4,
                z=original,
                discard if not={day}{4}
            ] {images/pcc-data.csv};
            \addplot3[
                area plot,
                fill=hgf-blue!17.14285714285714!white,
            ] table[
                col sep=comma, 
                x=dstep, 
                y expr=5,
                z=original,
                discard if not={day}{5}
            ] {images/pcc-data.csv};
            \addplot3[
                area plot,
                fill=hgf-blue!10!white,
            ] table[
                col sep=comma, 
                x=dstep, 
                y expr=6,
                z=original,
                discard if not={day}{6}
            ] {images/pcc-data.csv};
        \end{axis}
    \end{scope}
\end{tikzpicture}\qquad\begin{tikzpicture}
    \sffamily
    \begin{axis}[
        width=0.34\linewidth,
        height=0.25\linewidth,
        axis x line=left,
        axis y line=left,
        xlabel=time,
        ylabel=value,
        label style={font=\tiny\bfseries},
        scaled y ticks=false,
        xtick={0, 24, 48, 72, 96, 120, 144},
        xticklabels=\empty,
        yticklabels=\empty
    ]
        \addplot[thick, color=hgf-red] table[col sep=comma, x=step, y=pslp] {images/pcc-data.csv};
        \addplot[thick, color=hgf-green] table[col sep=comma, x=step, y=residual] {images/pcc-data.csv};
    \end{axis}
\end{tikzpicture}
    \caption{The concepts of \textit{primary cycle compression }(PCC) and \textit{learning residuals}. First, the original univariate time series (left) is rearranged according to its primary cycles, yielding a 2D data matrix (middle). Due to the similarity in primary cycles, we can compute recent history profiles (\localprofile{}) and subtract them from the original data, resulting in residuals that the model is trained to learn (right).}
    \label{fig:pcc}
\end{figure*}
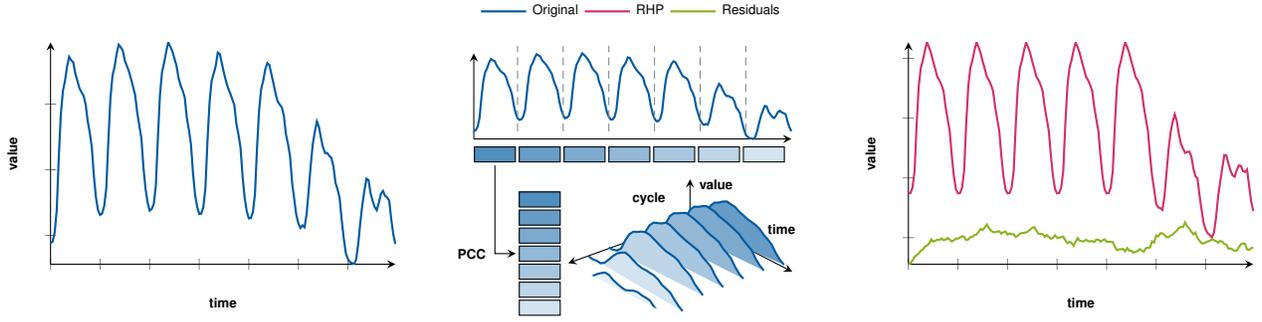

Let $x(t)$ be a time-dependent, continuous variable with $\featuredimension$ features at a point in time $t$. 
A time series $\mathbf{X}$ is a sequence of $N$ measurements of $x$ over a time span $T$, taken at times $t_0, t_1, ..., t_N$ with a temporal resolution of $\Delta t=T/N$.
We consider the time series forecasting problem as finding a mapping $\mathcal{M}$, such that 
\begin{equation*}
    \mathcal{M}\left(x(t_{i-\historicwindow}), \dots, x(t_{i-1}), x(t_i)\right) \rightarrow x(t_{i+1}, \dots, t_{i+\forecastwindow})
\end{equation*}
for every $x(t_i)~\in~\mathbf{X}$.
For simplicity, we will abbreviate $x(t_i)~:=~x_i$ moving forward.
$\historicwindow$ is the \textit{historic window length} and $\forecastwindow$ is the \textit{forecast window length}.

\section{Scalar Breakdown of Dot-Product Attention}

The success of the canonical Transformer is founded in the capabilities to relate individual sequence elements with one another through the dot-product self-attention mechanism spanning the attention matrix.
However, we conjecture that this does not hold for single-valued sequence elements, i.e. univariate time series.
Thus, for single-feature sequences, the attention matrix does not contain meaningful similarity information.
We refer to this as the \textit{scalar breakdown of dot-product attention}.

Let $\mathbf{X}$ be a sequences with feature dimension $\featuredimension=1$, $\sigma$ be an activation function, and $\mathbf{Q}$ and $\mathbf{K}$ be the matrices used to map to query and key respectively.
Then, the attention matrix element $a_{ij}$ associated with two arbitrary sequence elements $x_i, x_j \in \mathbf{X}$, before the softmax, is calculated as
\begin{equation}
    a_{ij} = \frac{\sigma\left(\mathbf{Q}x_i\right)\sigma\left(\mathbf{K}x_j\right)^T}{\sqrt{d_k}} \quad ,
\end{equation}
where $d_k$ is the dimension of the query and key vectors.
Since $d_k$ is only introduced to keep the results in a value range where the gradient of softmax is sufficient for effective backpropagation, it can be disregarded.
Yet, it also indicates that the values of query and key will generally be small.
In this range most common activation functions, excluding ReLU, can be approximated as linear, and since $x_i, x_j$ are scalar we obtain
\begin{equation}
\begin{aligned}
    a_{ij} &\propto \sigma\left(\mathbf{Q}x_i\right)\sigma\left(\mathbf{K}x_j\right)^T &&\approx x_i x_j\sigma\left(\mathbf{Q}\right)\sigma\left(\mathbf{K}\right)^T\\
    &\propto x_i x_j \left(\mathbf{Q}^* \mathbf{K}^{*T}\right) &&\propto x_i x_j \quad .
\end{aligned}
\label{eq:att_decomp}
\end{equation}
Hence, we argue that the product of two scalars does not contain meaningful information about their similarity as intended by the approach.

As mentioned, the ReLU activation function cannot be approximated linearly around zero.
However, it is explicitly linear on $\mathbb{R}^+$ and zero on $\mathbb{R}^-$.
Therefore, the above calculation can be performed similarly by considering the four possible sign combinations of $x_i$ and $x_j$, resulting in four different proportionality factors but leading to the same conclusion.

In the argument above the weight-biases $\mathbf{b}_Q, \mathbf{b}_K$ were neglected for clarity.
However, they can be easily reinserted into Eq.~\ref{eq:att_decomp}:
\begin{equation}
\begin{aligned}
    a_{ij} \propto & \,\sigma\left(\mathbf{Q}x_i + \mathbf{b}_Q\right)\sigma\left(\mathbf{K}x_j + \mathbf{b}_K\right)^T\\
    \approx & \,x_i x_j\sigma\left(\mathbf{Q}\right)\sigma\left(\mathbf{K}^T\right) + x_i \sigma\left(\mathbf{Q}\right) \sigma\left(\mathbf{b}_K\right)\\
    & + x_j \sigma\left(\mathbf{K}\right) \sigma\left(\mathbf{b}_Q\right) + \sigma\left(\mathbf{b}_Q\right) \sigma\left(\mathbf{b}_Q\right) \quad .
\end{aligned}
\end{equation}
None of the three additional terms contain both $x_i$ and $x_j$ and therefore cannot contribute to the similarity measure.

We hypothesize, that the authors of previous Transformer architectures for time series forecasting were implicitly aware of problems in applying dot-product attention to univariate time series, hence enriching the data through feature enhancement and positional encoding~\cite{wu_autoformer_nodate,zhou_informer_2021}, altering the attention mechanism e.g. convolutional attention~\cite{li2019enhancing} or simply focusing on multivariate time series. 
However, to the best of our knowledge, the deduction above is the first explicit reasoning for the scalar breakdown of dot-product attention, and it motivates our introducing the primary cycle compression.

\section{Methodology}
\label{sec:method}
To tackle the scalar break-down of dot-product attention for univariate time-series and to bring computational demand, runtime and energy consumption of Transformer-based architectures for long-term forecasting to a feasible level while improving forecast accuracy, we propose \fancyname{}, which is based on two novel concepts. 

\subsection{Primary Cycle Compression (PCC)}
\label{sec:pcc}

\begin{figure*}[ht]
    \centering
    \begin{tikzpicture}[node distance=\vspacing and \hspacing]
    % font style
    \sffamily

    % arrow styles
    \tikzstyle{arrow} = [->, -latex]
    \tikzstyle{rarrow} = [<-, -latex]
    
    % Adjustable numbers
    \def\spacing{1.0cm}
    \def\cwidth{2.5cm}
    \def\boxratio{0.8}
    
    % Variable names
    \def\datalength{L}
    \def\daylength{D}
    \def\metafeatures{M}
    \def\daycats{K}
    \def\forecastwindow{F}
    \def\historicwindow{H}
    \def\batchsize{B}
    
    % These are identical = \spacing by default but can be changed
    \def\hspacing{\spacing}
    \def\vspacing{\spacing/2}
    \def\lspacing{2}
    
    \def\rheight{\boxratio*\cwidth}
    \def\radius{.05*\spacing}
    
    \tikzstyle{box}=[
        rectangle, 
        fill=hgf-blue!10!white,
        minimum width=\cwidth, 
        minimum height=\rheight, 
        inner sep=0, 
        draw=black,
        font=\scriptsize\bfseries,
        rounded corners=1mm
    ]
    \tikzstyle{sbox}=[
        rectangle, 
        fill=hgf-green!10!white,
        minimum width=\cwidth, 
        minimum height=\rheight, 
        inner sep=0, 
        draw=black,
        font=\scriptsize\bfseries,
        rounded corners=1mm
    ]
    \tikzstyle{tshape}=[
        anchor=south, 
        inner sep=\lspacing,
        font=\tiny
    ]
    \tikzstyle{blabel}=[
        anchor=north, 
        inner sep=2*\lspacing,
        font=\scriptsize\bfseries
    ]
    \tikzstyle{elabel}=[
        midway,
        right=0.3em,
        inner sep=0,
        font=\tiny
    ]
    \tikzstyle{operator}=[
        circle,
        draw,
        inner sep=2,
        font=\tiny\bfseries
    ]
    \pgfplotsset{
        malformer/.style={
            width=1.3*\cwidth,
            height=1.3*\rheight,
            ticks=none,
            axis x line=center,
            axis y line=left,
            xmin=-10*pi,
            xmax=110*pi
        }
    }
    
    % Raw Data
    \node (raw_data) [box] {};
    \begin{scope}[shift={(raw_data)}, xshift=-0.3*\cwidth, yshift=-0.3*\rheight]
        \begin{axis}[malformer, ymin=0]
            \pgfmathsetseed{42};
            \addplot[black, samples=100, domain=0:100*pi] {3*sin(2*x)+2.5*sin(4*x)+20+2*rand};
        \end{axis};
    \end{scope};
    \node[blabel] at (raw_data.north) {Raw Data};
    \node[tshape] at (raw_data.south) {(\datalength$\cdot$\daylength)$\times$2};

    % Normalization Operator
    \node (norm_op) [operator, above right=0.5 of raw_data] {N};

    % Normalized
    \node (norm) [box, right=of raw_data, yshift=(.5*\hspacing+\rheight)] {};
    \begin{scope}[shift={(norm)}, xshift=-0.3*\cwidth, yshift=-0.3*\rheight]
        \begin{axis}[malformer, ytick={0,1}, yticklabels={0,1}]
            \pgfmathsetseed{42};
            \addplot[black, samples=100, domain=0:100*pi] {3/3*sin(2*x)+2.5/3*sin(4*x)+2/3*rand};
        \end{axis};
    \end{scope};
    \node[blabel] at (norm.north) {Normalized};
    \node[tshape] at (norm.south) {\datalength$\times$\daylength};

    % PSLP
    \node[box, right=of norm] (pslp) {};
    \begin{scope}[shift={(pslp)}, xshift=-0.3*\cwidth, yshift=-0.3*\rheight]
        \begin{axis}[malformer]
            \pgfmathsetseed{42};
            \addplot[black, samples=100, domain=0:100*pi] {3*sin(2*x)+2.5*sin(4*x)};
        \end{axis};
    \end{scope};
    \node[blabel] at (pslp.north) {Historic Profile};
    \node[tshape] at (pslp.south) {\datalength$\times$\daylength};

    % Meta Information
    \node[box, below=of pslp]  (meta) {$
        \begin{bmatrix}
            0 & 0 & 1\\
            0 & 1 & 0\\
            1 & 0 & 0\\
        \end{bmatrix}
    $};
    \node[blabel] at (meta.north) {Meta Information};
    \node[tshape] at (meta.south) {\datalength$\times$\metafeatures};

    % Residuals
    \node[box, right=of pslp]  (res) {};
    \begin{scope}[shift={(res)}, xshift=-0.3*\cwidth, yshift=-0.3*\rheight]
        \begin{axis}[malformer]
            \pgfmathsetseed{42};
            \addplot[black, samples=50, domain=0:100*pi] {2*rand};
        \end{axis};
    \end{scope};
    \node[blabel] at (res.north) {Residuals};
    \node[tshape] at (res.south) {\batchsize$\times$\forecastwindow$\times$\daylength};

    % Encoder
    \node[sbox, right=of res]  (enc) {};
    \begin{scope}[shift={(enc)}, xshift=-0.2*\cwidth]
        \draw[black, fill=hgf-green!30!white] (0, 0.2*\rheight) -- (0.4*\cwidth, 0.1*\rheight) -- (0.4*\cwidth, -0.1*\rheight) -- (0.0, -0.2*\rheight) -- cycle;
    \end{scope}
    \node[blabel] at (enc.north) {Encoder};
    \node[tshape] at (enc.south) {\batchsize$\times$\historicwindow$\times$(\daylength+\metafeatures)};

    % Decoder 
    \node[sbox, below=of enc] (dec) {};
    \begin{scope}[shift={(dec)}, xshift=-0.2*\cwidth]
        \draw[black, fill=hgf-green!30!white] (0, 0.1*\rheight) -- (0.4*\cwidth, 0.2*\rheight) -- (0.4*\cwidth, -0.2*\rheight) -- (0.0, -0.1*\rheight) -- cycle;
    \end{scope}
    \node[blabel] at (dec.north) {Decoder};
    \node[tshape] at (dec.south) {\batchsize$\times$\forecastwindow$\times$(\daylength+\metafeatures)};

    % Arrows
    \draw[arrow] (raw_data)[yshift=\vspacing] -| (norm_op);
    \draw[arrow] (norm_op) |- (norm);
    \draw[arrow] (raw_data) -- (meta);
    \draw[arrow] (norm) -- (pslp);
    \draw[arrow] (meta) -- (pslp);
    \draw[arrow] (pslp) -- (res);
    \draw[arrow] (res) -- (enc);
    \draw[arrow] (enc) -- (dec) node[elabel] {Latent repr.};

    \draw[arrow] ([xshift=0.5*\vspacing]res.south) |- (dec.west);
    \draw[arrow] (meta.east) -| ([xshift=-0.5*\vspacing]res.south);

\end{tikzpicture}
    \caption{Schematic overview of the data flow in \fancyname{}. Boxes represent building blocks, edges information flow, and tensor shapes are denoted at the bottom of each box.}
    \label{fig:flow}
\end{figure*}
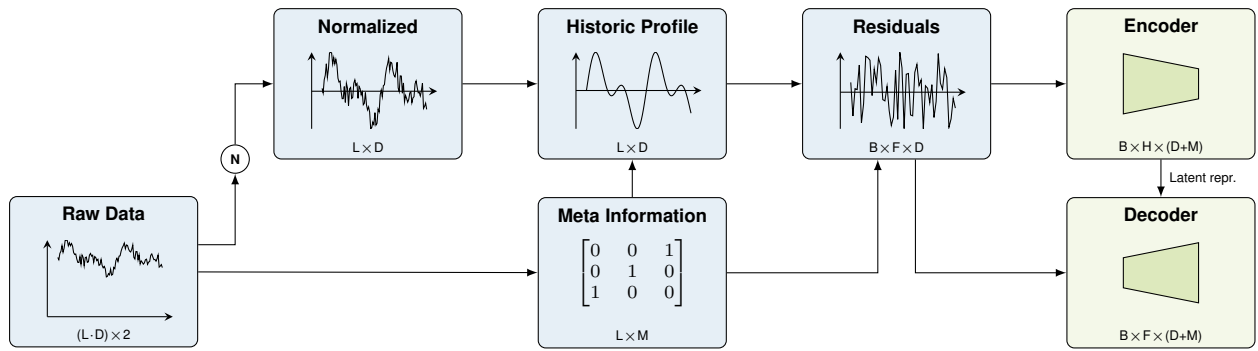

Transformer architectures for time series forecasting usually employ a sliding window that is applied to every time step.
While this creates a flexible model with respect to the starting point of the forecast, it also introduces two problems. 
Firstly, it causes data redundancy: depending on the length of historic and forecast windows a data point will be used tens or hundreds of times in each model training pass, i.e. epoch.
Secondly, Transformers work by determining the similarity of sequence elements by comparing elements primarily based on their absolute magnitude.
In time series forecasting, it is more relevant to know whether a sequence element is located in an ascending or descending slope, a peak, or a minimum.

We address both these shortcomings using what we coin \textit{primary cycle compression} (PCC).
The reason behind this is as simple as it is intuitive: in many time series one can observe periodicity in the data, a pronounced and stable \textit{primary cycle} (see Fig.~\ref{fig:pcc}).
This cycle is often, but not necessarily, the day-night cycle. 
For example, road traffic or demand for resource, such as electricity or water, align with daily activity of the population and industry, and hence exhibit a clear \SI{1}{\hour} cycle. Similar daily profiles can be observed in certain climate or weather data, like temperature or solar irradiation. 
Our approach uses this primary cycle as sequence granularity, expressing each step in the form of a vector with $\daylength{}$ entries, where $\daylength{}$ is the primary cycle length.
PCC enforces a fixed cycle start, however in practical applications long term forecasts are typically not performed every time step, but at the scale of the primary cycle.
Furthermore, it allows the dot product attention to compare the overall shapes of daily profiles, providing more relevant similarity measures.
Additionally, PCC reduces both sequence length, $\genericLength{}$, and number of samples by a factor of $\daylength{}$ while still utilizing the entire data content, thus mitigating many practical limitations originating from the $\mathcal{O}(\genericLength{}^2)$ memory and computational complexity of the attention mechanism.
Finally, PCC allows for natural inclusion of metadata \metafeatures{} available only on a whole-day basis, e.g. holiday vs. non-holiday; daily minimums, averages, or maximums; weather forecasts; etc.

\subsection{Recent Historic Profiles and Residuals}

PCC allows us to incorporate knowledge about predominating patterns in the data, focusing the model on learning more difficult to capture temporal dependencies beyond the trivial periodic patterns, as illustrated in Figure \ref{fig:pcc}.

Recurring patterns may be captured by averaging several past instances of the data, resulting in so-called \textit{historic profiles} (\historicprofile).
There are different ways for averaging historic data and typically each application use-case necessitates a custom combination of measurements.
The simplest form of a historic profile is \textit{persistence}, where data from the last time step $i$ is used to forecast the next one, $i+1$.
Following empirical evaluation of different averaging techniques, we identify \textit{recent history profiles} (\localprofile{}) as an innovative and precise technique for a wide range of use cases.
However, all further model derivations can be easily translated to other averaging techniques.
For calculation of \localprofile{} we distinguish between \daycats{} different types, which enabled incorporation of prior knowledge. A typical example for different categories would be workdays, saturdays and sun-/holidays, since many datasets exhibit distinct patterns in these categories.
Instead of considering all past data we only consider the last $\pslpwindow{}$ days in the appropriate category. That way, the profiles naturally account for current circumstances (like weather), seasonality, behavior change and concept drift.

Historic profiles capture the general time course of the primary cycle rather well, but smooth out faster, more erratic components.
To let the model focus on these hard-to-predict modulations rather than the already known predominant general time course, we propose a residual forecasting approach, were input and target of the network are the difference between the \localprofile{} and the actual time curve, i.e. the \textit{residual} (see Fig.~\ref{fig:pcc}).
Since the overall magnitude of the time series influences the forecast of residuals, we provide the \localprofile{} as decoder input.
In addition to providing a baseline, this also means the decoder input is shorter than in previous approaches~\cite{zhou_informer_2021}, resulting in a trivial reduction of computations.
With the residual approach, the network can easily learn to return zero if it is unable to find further patterns on top of the \localprofile{}, ensuring a robust fallback behavior of the prediction.

\subsection{ReCycle}

Our approach \fancyname{} combines PCC and residual forecasting based on RHP into one stand-alone preprocessing step that can be applied to any kind of Transformer-based model architecture, and in fact any time-series forecasting model. \fancyname{} works as follows:
Time-series data is normalized to the interval $[0,1]$.
PCC is then performed on the whole dataset, i.e. a univariate time series with $N$ time steps is rearranged into a 2D data matrix of size $\datalength{}\times\daylength{}$, with $\datalength{}=N/\daylength$ and $\daylength$ being the length of the primary cycle. $\daylength{}$ is an application-specific parameter. 
Metadata, such as weekday and holiday information extracted from time stamps of each sample, is then concatenated as additional features.
\localprofile{}s are calculated for each sample in the dataset, regarding the last $\pslpwindow{}$ days of category \daycats. Residuals are then determined by subtracting the respective \localprofile{} from the original data for each time step. 

The resulting dataset is then fed to the actual model, which can be any kind of encoder-decoder architecture for sequence-to-sequence modeling. We will elucidate the process on the example of the vanilla Transformer architecture, depicted in Fig.~\ref{fig:flow}:
The model takes the sequence of historic residuals of size $\historicwindow{}\times(\daylength{}+\metafeatures{})$ as input, where $\historicwindow{}$ is the length of the historic window. 
Residuals are encoded through one multi-head attention layer into the hidden state and passed to the decoder.
The decoder takes the hidden state and the forecast \localprofile{} of size $\forecastwindow{}\times(\daylength{}+\metafeatures{})$ as input and outputs forecast residuals.

Since the output length of the prediction sequence is known a-priori in time series forecasting, we use \textit{single pass decoding} (SPD) in both training and inference, i.e. predicting all sequence elements at once instead of one sequence element after another, as is common for Transformers in NLP.
This approach has been widely adopted since its introduction in \cite{zhou_informer_2021}.
\decoding{} comes with two advantages: For one, it reduces computational complexity, since it allows leveraging the natural vector processing capabilities of the attention mechanism. 
It further grants better information utilization~\cite{wu_autoformer_nodate, zhou_fedformer_2022} and provides a significant contribution to performance improvements \cite{zeng_are_2022}.
We also use it to forego masking, allowing the model to use the known \localprofile{} predictions of future time steps. 
%In empirical studies we found that, counter-intuitively, \decoding{} actually improves prediction accuracy.
%We hypothesize that it prevents error accumulation in longer sequences and hence leads to more accurate forecasts.

\begin{figure*}[!ht]
    \centering
    \begin{tikzpicture}
    \sffamily
    \begin{groupplot}[
        group style={
            group name=result plot,
            group size=2 by 2,
            xlabels at=edge bottom,
            ylabels at=edge left,
            horizontal sep=1.5cm,
            vertical sep=0.8cm
        },
        width=0.45\linewidth,
        height=0.25\linewidth,
        axis x line=left,
        axis y line=left,
        % ylabel=residual load {[GW]},
        label style={font=\tiny\bfseries},
        tick label style={font=\tiny\sansmath\sffamily},
        xtick={0, 24, 48, 72, 96, 120, 144},
        xticklabels={0, 1, 2, 3, 4, 5, 6},
        legend columns=4,
        legend cell align={left},
        legend style={
            draw=none,
            font=\tiny,
            at={(1.0, 2.3)},
            anchor=south
        }
    ]
    \nextgroupplot[ylabel=residual load {[GW]}]
        \addplot[thick, color=hgf-gray!50!white, densely dotted] table[col sep=comma, x=step, y=reference] {images/entsoe-res.csv};
        \addplot[thick, color=hgf-blue] table[col sep=comma, x=step, y=residual] {images/entsoe-res.csv};
    \nextgroupplot[ylabel=residual water usage {[m\textsuperscript{3}]}]
        \addplot[thick, color=hgf-gray!50!white, densely dotted] table[col sep=comma, x=step, y=reference] {images/water-res.csv};
        \addplot[thick, color=hgf-blue] table[col sep=comma, x=step, y=residual] {images/water-res.csv};
    \nextgroupplot[xlabel=time {[d]}, ylabel=load {[GW]}]
        \addplot[thick, color=hgf-red] table[col sep=comma, x=step, y=Transformer] {images/entsoe-comp.csv};
        \addlegendentry{Transformer};
        \addplot[thick, color=hgf-blue] table[col sep=comma, x=step, y=ReCycle] {images/entsoe-comp.csv};
        \addlegendentry{Transformer + ReCycle};
        \addplot[thick, color=hgf-gray!50!white, densely dotted] table[col sep=comma, x=step, y=Target] {images/entsoe-comp.csv};
        \addlegendentry{Target};
    \nextgroupplot[xlabel=time {[d]}, ylabel=water usage {[m\textsuperscript{3}]}]
        \addplot[thick, color=hgf-red] table[col sep=comma, x=step, y=Transformer] {images/water-comp.csv};
        \addplot[thick, color=hgf-blue] table[col sep=comma, x=step, y=ReCycle] {images/water-comp.csv};
        \addplot[thick, color=hgf-gray!50!white, densely dotted] table[col sep=comma, x=step, y=Target] {images/water-comp.csv};
    \end{groupplot}
\end{tikzpicture}
    \caption{Exemplary plots of target and predicted residuals (top) and full sample (bottom), for the two datasets \entsoe{} (left) and \water{} (right).}
    \label{fig:result_plot}
\end{figure*}
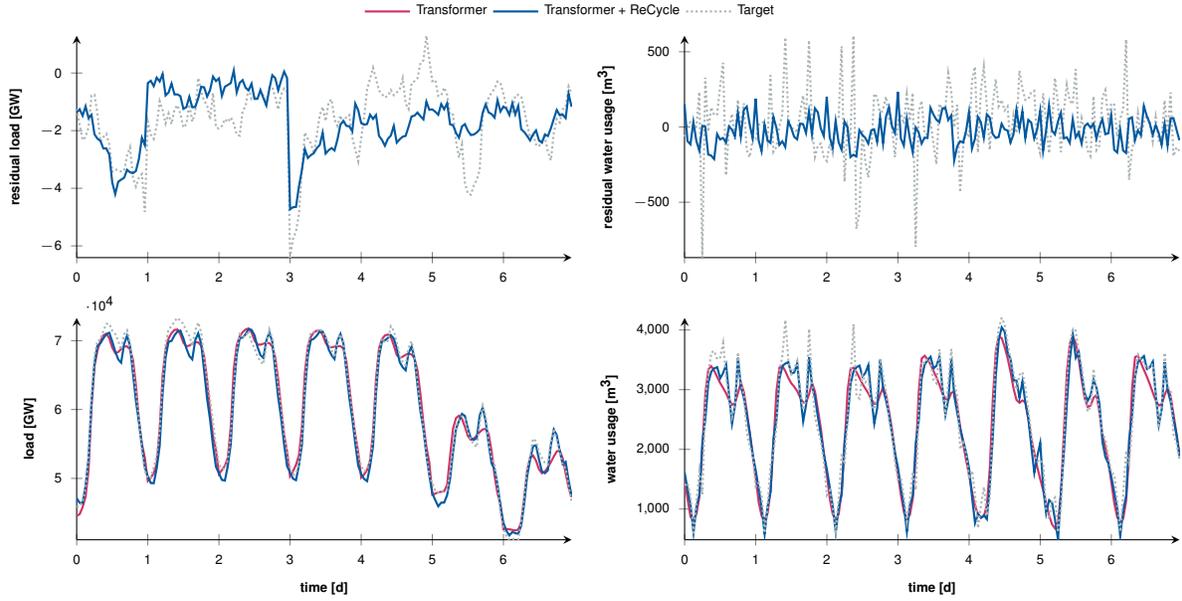

\section{Experimental Evaluation}
We evaluate \fancyname{} as an extension to three current SOTA Transformer-based models on a number of time series forecasting problems. Results were compared in terms of prediction accuracy, training time and energy consumption for training and inference.

\subsection{Benchmarks}
The evaluation task is defined as predicting values of the next $\forecastwindow=7$ days (168 hours), based on data from the last $\historicwindow=21$ days (504 hours). %, i.e. $\historicwindow=21$ and $\forecastwindow=7$.
Three representative Transformer-based architectures were chosen: the vanilla Transformer, FEDformer and PatchTST. While originally developed for NLP sequence prediction, the \textit{Vanilla Transformer} as described by \cite{vaswani_attention_2017} can be used to directly encode historic data and forecast future temporal behavior.
\textit{FEDformer}~\cite{zhou_fedformer_2022} and \textit{PatchTST} mark current state-of-the-art Transformer variants for long time series forecasting.
For Transformer and FEDformer we use the implementations provided by~\cite{zhou_fedformer_2022}\footnote{https://github.com/MAZiqing/FEDformer}. For PatchTST we use the original implementation provided by the authors\footnote{https://github.com/yuqinie98/PatchTST}.
All three Transformer-based models were trained with and without ReCycle.

We further include the NHiTS model~\cite{challu2023nhits} into our experimental evaluation. NHiTS is a non-Transformer approach that utilizes hierarchical interpolation and multi-rate data sampling to improve prediction accuracy and reduce training time. We use the implementation provided by the authors~\footnote{https://github.com/cchallu/n-hits} and train it with the standard setup.
For reference, we report predictive performance of the \localprofile{} we use to calculate residuals. %Dedicated ablation studies of the individual components of \fancyname{} are included in the appendix. 
\paragraph{Hyperparameters}
Hyperparameters of the FEDformer, PatchTST and NHiTS were set according to the original publications. Since Transformer was originally designed for sequence processing, we conducted a hyperparameter-search using Propulate~\cite{taubert2023propulate}. 
Training with and without \fancyname{} was conducted using the same set of hyperparameters for each model.
\paragraph{Setup}
For application of \fancyname{}, PCC is performed as described above. We consider the primary cycle to be the daily one, i.e. we fix $\daylength{}=$ \SI{24}{\hour}.
Recent historic profiles are extracted by categorizing each sample into one of $\daycats{}=3$ categories: Weekday; Saturday; Sun- or Holiday.
The \localprofile{} for each sample is then calculated by averaging the last $\pslpwindow = 3$ samples prior to the sample date in a given category. This implies that at least three weeks of data have to be available before a prediction can be made.
Furthermore we use weekday and holiday information, extracted from time stamps of each as metadata features, such that weekday information is one-hot encoded and holiday information for the current and following day is binary encoded. Thus, there is a total of $\metafeatures{}=9$ metadata features.
Models are then trained using the Adam optimizer, without an additional learning rate schedule. Since \fancyname{} is trained to predict residuals, which intrinsically are more noisy compared to the original data, we optimize for mean absolute error (MAE) loss, which provides higher resilience to outliers than mean square error (MSE) loss.

\subsection{Datasets}
Five representative time series from different datasets were chosen for experimental evaluation. %Details can be found in the Appendix.
\textit{\textbf{Electricity Load Diagrams}}\footnote{\noindent\url{https://archive.ics.uci.edu/ml/datasets/ElectricityLoadDiagrams20112014}} (\ucipt) contains electricity consumption of 370 household consumers in Portugal, taken between January 2011 and December 2014 at a temporal resolution of \SI{15}{\min}.
Following the approach first used in \cite{zhou_informer_2021} for univariate forecasting we only use the column 'MT\_320'.
%\minigrid\footnote{This dataset was provided by DLR and will be made publicly available upon publication.} contains time-resolved measurements of electricity consumption of a rural minigrid installation in Tanzania, taken between March 2020 to March 2021 at a temporal resolution of \SI{15}{\min}. 
%The minigrid system comprises a photovoltaic solar power installation, coupled with energy storage, i.e. batteries, and a fossil fuel generators. Precise predictions of electricity consumption in the grid can aid in optimal scheduling of generator operating hours, determining charge cycles of the energy storage device and performing predictive maintenance.\\
\textit{\textbf{Western European Power Consumption}}\footnote{\url{https://www.kaggle.com/datasets/francoisraucent/western-europe-power-consumption}} (\entsoede) contains time-resolved measurements of total electricity consumption of Germany, collected by the ENTSO-E, taken between January 2015 and August 2020 at a temporal resolution of \SI{15}{\min}.
\textit{\textbf{Water Demand}}\footnote{This dataset was provided by Siemens AG and can be found at\\\url{https://doi.org/10.5281/zenodo.11045013}} (\water) contains time-resolved measurements of water consumption in a water supply network, taken between January 2016 and March 2021 at a temporal resolution of \SI{1}{\hour}.
Measurements of water consumption were collected at a pumping station in a regional water supply network in Germany for both household and industrial consumers.
\textit{\textbf{Traffic}}\footnote{\url{https://pems.dot.ca.gov/?dnode=Freeway&content=loops&tab=det_timeseries&fwy=280&dir=N}} (\traffic) contains time-resolved measurements of water consumption in a water supply network, taken between January 2016 and March 2021 at a temporal resolution of \SI{1}{\hour}.  It can be found on \url{https://pems.dot.ca.gov} under Freeways $\rightarrow$ I280-N $\rightarrow$ Performance $\rightarrow$ Aggregates $\rightarrow$ Time Series.
\textit{\textbf{Electricity Transformer Temperature}} (\etth)\footnote{\url{https://github.com/zhouhaoyi/ETDataset}} contains measurements of power load features and oil temperature of two electricity transformers in China, taken between July 2016 and June 2018. We use the dataset of the second transformer with a temporal resolution of \SI{1}{\hour} (\etthtwo{}).
%\textit{Electricity Load Diagrams} (\ucipt) contains electricity consumption of 370 household consumers in Portugal, taken between January 2011 and December 2014 at a temporal resolution of \SI{15}{\min}.
%\textit{Western European Power Consumption} (\entsoede) contains time-resolved measurements of total electricity consumption of Germany, collected by the ENTSO-E, taken between January 2015 and August 2020 at a temporal resolution of \SI{15}{\min}. 
%\textit{Water Demand} (\water) contains time-resolved measurements of water consumption in a water supply network, taken between January 2016 and March 2021 at a temporal resolution of \SI{1}{\hour}.
%\textit{Traffic} contains Vehicle Miles Traveled on the I280-N from San Jose to San Francisco taken between January 2018 to December 2022 at a temporal resolution of \SI{1}{\hour}.
%\textit{Electricity Transformer Temperature} (\etth{}) contains measurements of power load features and oil temperature of two electricity transformers in China, taken between July 2016 and June 2018. We use the dataset of the second transformer with a temporal resolution of \SI{1}{\hour} (\etthtwo{}).
%Datasets with a temporal resolution higher than one hour are sampled down to one hour by averaging all data points in an hour.
Where necessary datasets are downsampled to \SI{1}{\hour}.
Missing values are filled in with the average of the two neighboring values since this trivially generalizes to any resolution.
Each dataset is separated into training, validation and test set in the ratio 6:2:2 along the temporal axis.

\subsection{Compute Infrastructure}
All models were run on a single node of a supercomputing cluster, equipped with two Intel Xeon ``Ice Lake'' processor cores and 4 NVIDIA A100 Tensor Core GPUs. 
The system allows for measurements of whole-node energy consumption via sensors of Lenovo's XClarity Controller (XCC), which can be read via IPMI and SLURM.
Models were implemented in Python 3.9.2 using the PyTorch framework~\cite{paszke2019pytorch} versioned 1.13.1+cu116 with CUDA version 11.6. Runs were performed using a single NVIDIA A100-40 GPU for each model.

\subsection{Results}
\begin{table}
\centering
\caption{Forecasting performance of Transformer-based models with and without usage of \fancyname{} on five datasets for a forecast window of one week. Best results are highlighted in italics. MAPE is provided in percent. MSE is rescaled per dataset for readability, the respective orders of magnitude are noted in square brackets next to the dataset.}
\small
\begin{tabular}{clrr}
\toprule
             & & MSE            & MAPE         \\
             \midrule
\parbox[t]{1mm}{\multirow{8}{*}{\rotatebox[origin=c]{90}{\entsoede{} [$10^6$]}}}&
Transformer                  & 8.53$\pm$0.31            & 3.59$\pm$0.14          \\
& \hspace{0.5em}+ \fancyname{} & \textbf{5.80$\pm$0.58} & \textbf{3.19$\pm$0.17} \\
& FEDformer                  & 11.5$\pm$0.14            & 4.03$\pm$0.03          \\
& \hspace{0.5em}+ \fancyname{} & 6.36$\pm$0.22          & 3.25$\pm$0.05          \\
& PatchTST                   & 14.5$\pm$0.13            & 4.54$\pm$0.02          \\
& \hspace{0.5em}+ \fancyname{} & 7.97$\pm$0.10          & 3.49$\pm$0.02          \\
& NHiTS                      & 7.90$\pm$0.01            & \textbf{3.19$\pm$0.01} \\
& \localprofile              & 8.85                     & 3.89                   \\
\midrule
\parbox[t]{2mm}{\multirow{8}{*}{\rotatebox[origin=c]{90}{\ucipt{} [$10^1$]}}}&
Transformer                  & 11.30$\pm$0.27           & 11.30$\pm$0.44         \\
& \hspace{0.5em}+ \fancyname{} & 9.60$\pm$0.29          & 6.25$\pm$0.27          \\
& FEDformer                  & \textbf{8.34$\pm$0.01}   & 9.14$\pm$0.03          \\
& \hspace{0.5em}+ \fancyname{} & 11.70$\pm$1.85         & 7.37$\pm$0.84          \\
& PatchTST                   & 7940.0$\pm$152.0         & \textbf{5.52$\pm$0.05} \\
& \hspace{0.5em}+ \fancyname{} & 8.88$\pm$0.04          & 5.71$\pm$0.01          \\
& NHiTS                      & 9.86$\pm$0.23            & 6.66$\pm$0.16          \\
& \localprofile              & 9.44                     & 6.52                   \\
\midrule
\parbox[t]{2mm}{\multirow{8}{*}{\rotatebox[origin=c]{90}{\water{} [$10^4$]}}}&
Transformer                  & 16.3$\pm$0.2             & 15.1$\pm$0.1           \\
& \hspace{0.5em}+ \fancyname{} & 17.4$\pm$0.6           & 14.3$\pm$0.1           \\
& FEDformer                  & \textbf{14.3$\pm$0.1}    & 15.6$\pm$0.1           \\
& \hspace{0.5em}+ \fancyname{} & 17.6$\pm$0.3           & 15.0$\pm$0.2           \\
& PatchTST                   & 27.9$\pm$1.4             & 19.8$\pm$0.4           \\
& \hspace{0.5em}+ \fancyname{} & 14.9$\pm$0.1           & \textbf{13.8$\pm$0.1}  \\
& NHiTS                      & 163.9$\pm$0.3            & 61.5$\pm$0.1           \\
& \localprofile              & 14.5                     & 13.9                   \\
\midrule
\parbox[t]{2mm}{\multirow{8}{*}{\rotatebox[origin=c]{90}{\traffic{} [$10^8$]}}}&
Transformer                    & 2.77$\pm$0.14          & 16.1$\pm$2.1           \\
& \hspace{0.5em}+ \fancyname{} & 1.43$\pm$0.04          & 10.7$\pm$1.4           \\
& FEDformer                    & 2.41$\pm$0.03          & 15.2$\pm$0.1           \\
& \hspace{0.5em}+ \fancyname{} & 1.28$\pm$0.08          & 9.0$\pm$0.9            \\
& PatchTST                     & 2.17$\pm$0.04          & 12.8$\pm$0.5           \\
& \hspace{0.5em}+ \fancyname{} & \textbf{1.23$\pm$0.01} & \textbf{7.3$\pm$0.1}   \\
& NHiTS                        & 2.46$\pm$0.01          & 9.3$\pm$0.2            \\
& \localprofile                & 2.02                   & 9.7                    \\
\midrule
\parbox[t]{2mm}{\multirow{8}{*}{\rotatebox[origin=c]{90}{\etthtwo{} [$10^1$]}}}&
Transformer                    & 4.00$\pm$0.25          & 37.1$\pm$1.1           \\
& \hspace{0.5em}+ \fancyname{} & 3.64$\pm$0.18          & 17.7$\pm$0.3           \\
& FEDformer                    & 4.11$\pm$0.04          & 28.4$\pm$1.5           \\
& \hspace{0.5em}+ \fancyname{} & 4.37$\pm$0.17          & 18.7$\pm$0.4           \\
& PatchTST                     & \textbf{2.15$\pm$0.01} & 139.0$\pm$0.6          \\
& \hspace{0.5em}+ \fancyname{} & 4.33$\pm$0.01          & 18.9$\pm$0.1           \\
& NHiTS                        & 4.39$\pm$0.05          & \textbf{17.3$\pm$0.1} \\
& \localprofile                & 5.03                   & 20.99                  \\
\bottomrule
\end{tabular}
\label{tab:results}
\end{table}

Table \ref{tab:results} summarizes the predictive performance of training runs of all studied approaches.
%Table \ref{tab:energy} summarizes wallclock time and energy consumption of the same runs.

We report mean square error (MSE) as well as mean absolute percentage Error (MAPE), which is the mean absolute error (MAE) normalized to the original value.
In addition to prediction accuracy metrics, we measure run time as well as total energy consumption of model training for model architectures, as they are known to be computationally complex and are thus energy-hungry.
Towards this end, each model is evaluated three times and the average time provided, while energy consumption is measured as the total consumption of all three evaluations.
%Table \ref{tab:energyresults} shows corresponding measurements for wallclock time and energy consumption of training runs.

Throughout all datasets, the naive approach of using plain \localprofile{} yields predictive performance competitive with the evaluated DL-models. Nonetheless, training a Transformer-based model architecture on \localprofile{} residuals always yields improved forecasting, demonstrating the benefit of these models of simple statistical methods.

\begin{table}[h!]
\centering
\caption{Forecast window length ablation study of Transformer-based models with and without usage of \fancyname{} on two datasets.}
\footnotesize
%\begin{tabular}{clrrrr}
\begin{tabular}{cl r@{\hskip 5pt} r@{\hskip 15pt} r@{\hskip 5pt} r }
\toprule
              
              & & Seq.    & Length = 96  & Seq.    & Length = 336    \\
              & & MSE            & MAPE         & MSE            & MAPE    \\
              \midrule
\parbox[t]{1mm}{\multirow{7}{*}{\rotatebox[origin=c]{90}{\entsoede{} [$10^6$]}}}&
Transformer                  & 7.86$\pm$0.44             & 3.32$\pm$0.11           & 8.92$\pm$0.07          & 3.63$\pm$0.04  \\
& \hspace{0.5em}+ \fancyname{}& 5.69$\pm$0.23            & 3.17$\pm$0.09           & 5.86$\pm$0.33          & 3.19$\pm$0.03  \\
& FEDformer                  & 9.85$\pm$0.11             & 6.69$\pm$0.01           & 13.2$\pm$0.67          & 4.44$\pm$0.18  \\
& \hspace{0.5em}+ \fancyname{} & 6.23$\pm$0.46           & 3.29$\pm$0.11           & 6.12$\pm$0.23          & 3.26$\pm$0.04  \\
& PatchTST                   & 13.6$\pm$0.46             & 4.34$\pm$0.06           & 13.9$\pm$.043          & 4.65$\pm$0.09  \\
& \hspace{0.5em}+ \fancyname{} & 7.97$\pm$0.13           & 3.50$\pm$0.02           & 8.03$\pm$0.08          & 3.51$\pm$0.02  \\
& NHiTS                      & 7.41$\pm$0.07             & 2.93$\pm$0.02           & 6.92$\pm$0.40          & 3.14$\pm$0.07  \\
\midrule
\parbox[t]{2mm}{\multirow{7}{*}{\rotatebox[origin=c]{90}{\etthtwo{} [$10^1$]}}}&
Transformer                  & 38.8$\pm$0.9             & 43.3$\pm$3.8             & 40.8$\pm$0.4          & 31.8$\pm$0.4  \\
& \hspace{0.5em}+ \fancyname{}& 40.5$\pm$3.2            & 18.3$\pm$0.1             & 38.4$\pm$3.3          & 18.0$\pm$0.5  \\
& FEDformer                  & 35.1$\pm$0.1             & 30.8$\pm$0.9             & 46.2$\pm$1.1          & 26.9$\pm$1.1  \\
& \hspace{0.5em}+ \fancyname{} & 44.7$\pm$0.6           & 19.1$\pm$0.1             & 46.7$\pm$2.5          & 19.4$\pm$0.5  \\
& PatchTST                   & 17.2$\pm$0.1             & 119.0$\pm$3.13           & 24.9$\pm$0.2          & 168.0$\pm$2.2 \\
& \hspace{0.5em}+ \fancyname{} & 43.5$\pm$0.1           & 18.8$\pm$0.1             & 43.4$\pm$0.4          & 18.8$\pm$0.1  \\
& NHiTS                      & 37.9$\pm$0.3             & 16.9$\pm$0.2             & 31.9$\pm$0.2          & 12.4$\pm$0.1  \\

\bottomrule
\end{tabular}
\label{tab:ablation}
\end{table}

Results show, that no clearly superior model can be identified and that in fact, depending on the datasets, different approaches yield the best prediction accuracy.
Furthermore, the model yielding lowest MSE differs from the model yielding lowest MAPE in almost all cases.
For the \entsoe{} and \traffic{} datasets, adding \fancyname{} improves both MSE and MAPE for all Transformer-based approaches.
For \water{} and \ucipt{} datasets, plain FEDformer yields lowest MSE values, but not lowest MAPE values. On those two datasets, adding \fancyname{} to the vanilla Transformer and PatchTST also improves MSE and MAPE. Further investigation showed, that these two datasets consist of very noisy time series. Hence, only few patterns remain in the residuals and the model detects almost exclusively noise, a illustrated in Fig.~\ref{fig:result_plot}.
In that case, \fancyname{} provides a reliable fallback behavior, where the model returns the \localprofile{} prediction, if it cannot identify recurring patterns. 
Thus, the simple nature of the \localprofile{} provides a worst case scenario that is both explainable and robust.
The \etthtwo{} datasets constitutes a special case, as application of \fancyname{} clearly improves MAPE values, but not MSE values. This is likely caused by the fact that models with \fancyname{} were trained on optimizing MAE, whereas the vanilla versions are trained on MSE loss.
Table \ref{tab:ablation} shows ablation with respect to forecast window length on two datasets, which demonstrates consistent behavior across for the different forecast window lengths.

The true benefit of using \fancyname{} becomes evident when considering training time and energy consumption, which is shown in Table~\ref{tab:energy}. Results clearly demonstrate, that \fancyname{} helps to drastically reduce both compute time and energy consumption for training Transformer-based approaches.
Notably, the reduction in energy consumption does not stem solely from shorter training times.
For the vanilla Transformer, using \fancyname{} achieves a training speed-up of three to five, while reducing energy consumption by a factor of up to $\approx 20$ (\water{} dataset).
The biggest improvements can be observed for FEDFormer, where usage of \fancyname{} yields up to $\approx 18$ times faster training (\etthtwo{} dataset) and up to $\approx 35$ times lower energy consumption (\ucipt{} dataset).
In those cases, where FEDformer yields the best forecasting accuracy in terms of MSE, the decrease of $\approx 30\%$ (\ucipt{}) and $\approx$ 14\% (\water{}) in MSE is to be gauged against a factor of $\approx 13 - 15$ longer training time and $\approx 22 - 35$ higher energy consumption.
The same holds for PatchTST in the \etthtwo{} dataset, where conceding a two times higher MSE can reduce training time by a factor of $\approx 10$ and energy consumption by a factor of $\approx 16$.
NHiTS yields low training times and energy consumption throughout all datasets, however, Transformer-based approaches with \fancyname{} provide even lower values. For the \entsoe{} and the \etthtwo{} datasets, the difference between the best MAPE achieved by NHiTS and the runner up Transformer + \fancyname{} is almost negligible, while the latter runs at $\approx$ 50-75\% shorter training and corresponding lower energy.

\begin{table}
\centering
\caption{Average run time, energy consumption and reduction factors achieved through \fancyname{} of Transformer-based models with and without usage of \fancyname{} on five datasets for a forecast window of one week. Lowest absolute values are highlighted in bold.}
\small
\begin{tabular}{clrrrr}
\toprule
             &                 & Time              &                   & Energy                 &                   \\
             &                 & [\SI{}{\second}]  &          & [\SI{}{\watt\hour}]    &          \\
             \midrule
\parbox[t]{1mm}{\multirow{7}{*}{\rotatebox[origin=c]{90}{\entsoede}}}&
Transformer                    & 161.09            &  \multirow{2}{*}{$\times 2.8$}          & 111.0      &  \multirow{2}{*}{$\times 9.0$}                 \\
& \hspace{0.5em}+ \fancyname{} & \textbf{56.96}    &                & \textbf{12.3}          &                \\
& FEDformer                    & 1242.81           & \multirow{2}{*}{$\times 10.1$}          & 723.0  & \multirow{2}{*}{$\times 23.8$}\\
& \hspace{0.5em}+ \fancyname{} & 123.61            &                   & 30.4                   &               \\
& PatchTST                     & 315.62            & \multirow{2}{*}{$\times 3.7$}& 164.0                  &\multirow{2}{*}{$\times 10.1$}\\
& \hspace{0.5em}+ \fancyname{} & 85.71             &                & 16.2                   &               \\
& NHiTS                        & 86.71             &                   & 19.0                   &                   \\
\midrule
\parbox[t]{2mm}{\multirow{7}{*}{\rotatebox[origin=c]{90}{\ucipt}}}&
Transformer                    & 105.99            &\multirow{2}{*}{$\times 4.5$}& 74.2                   &\multirow{2}{*}{$\times 12.5$}\\
& \hspace{0.5em}+ \fancyname{} & \textbf{23.62}    &                & \textbf{5.92}          &               \\
& FEDformer                    & 977.55            &\multirow{2}{*}{$\times 15.4$}& 560.0    &\multirow{2}{*}{$\times 34.4$}\\
& \hspace{0.5em}+ \fancyname{} & 63.30             &               & 16.3               &               \\
& PatchTST                     & 221.05            &\multirow{2}{*}{$\times 4.4$}& 115.0    &\multirow{2}{*}{$\times 9.8$}\\
& \hspace{0.5em}+ \fancyname{} & 50.64             &                & 11.7                   &                \\
& NHiTS                        & 81.98             &                   & 17.9                   &                   \\
\midrule
\parbox[t]{2mm}{\multirow{7}{*}{\rotatebox[origin=c]{90}{\water}}}&
Transformer                    & 250.97            &\multirow{2}{*}{$\times 5.6$}& 169.0                  &\multirow{2}{*}{$\times 17.8$}\\
& \hspace{0.5em}+ \fancyname{} & 44.71             &                & \textbf{9.5}           &               \\
& FEDformer                    & 1243.55           &\multirow{2}{*}{$\times 13.3$}& 712.0    & \multirow{2}{*}{$\times 32.2$}\\
& \hspace{0.5em}+ \fancyname{} & 93.34             &               & 22.1                   &               \\
& PatchTST                     & 331.55            &\multirow{2}{*}{$\times 7.5$}& 173.0    & \multirow{2}{*}{$\times 16.0$}\\
& \hspace{0.5em}+ \fancyname{} & \textbf{44.23}    &                & 10.8                   &               \\
& NHiTS                        & 85.8              &                   & 20.4                   &                   \\
\midrule
\parbox[t]{2mm}{\multirow{7}{*}{\rotatebox[origin=c]{90}{\traffic}}}&
Transformer                    & 111.54            &\multirow{2}{*}{$\times 2.9$}& 79.0 &\multirow{2}{*}{$\times 6.4$}\\
& \hspace{0.5em}+ \fancyname{} & \textbf{38.91}    &                & 12.3                   &                \\
& FEDformer                    & 1108.96           &\multirow{2}{*}{$\times 10.3$}& 646.0    & \multirow{2}{*}{$\times 24.6$}\\
& \hspace{0.5em}+ \fancyname{} & 107.56            &               & 26.3                   &               \\
& PatchTST                     & 319.91            &\multirow{2}{*}{$\times 7.9$}& 164.0    & \multirow{2}{*}{$\times 19.8$}\\
& \hspace{0.5em}+ \fancyname{} & 40.33             &                & \textbf{8.28}          &               \\
& NHiTS                        & 86.61             &                   & 18.7                   &                   \\
\midrule
\parbox[t]{2mm}{\multirow{7}{*}{\rotatebox[origin=c]{90}{\etthtwo}}}&
Transformer                    & 73.11             &\multirow{2}{*}{$\times 5.0 $} & 50.4    & \multirow{2}{*}{$\times 12.3$}\\
& \hspace{0.5em}+ \fancyname{} & \textbf{14.54}    &               & \textbf{4.1}           &               \\
& FEDformer                    & 369.74            & \multirow{2}{*}{$\times 17.9$}& 216.0   & \multirow{2}{*}{$\times 28.8$}\\
& \hspace{0.5em}+ \fancyname{} & 20.71             &               & 7.5                    &               \\
& PatchTST                     & 152.41            & \multirow{2}{*}{$\times 9.6$}& 80.6    & \multirow{2}{*}{$\times 16.4$}\\
& \hspace{0.5em}+ \fancyname{} & 15.87             &                & 4.9                    &               \\
& NHiTS                        & 77.65             &                   & 16.1                   &                   \\
\bottomrule
\end{tabular}
\label{tab:energy}
\end{table}

\section{Conclusion}
\label{sec:discussion}
Attributing to its central role in many real world applications, time series forecasting has received a lot of attention in recent years. A plethora of approaches have been published, many of them relying of the Transformer architecture. However, in the thrive for ever better prediction accuracy, the demand in computational resources of current models has far outgrown anything feasible for practical deployment. The steadily increasing energy consumption required to (re\nobreakdash)train and run these models not only limits their application e.g. on embedded or edge devices, it also constitutes a further step along the unsustainable path that AI research is currently on.
In response to that, we present \fancyname{}, a method for reducing runtime and energy consumption in long time series forecasting with Transformer-based architectures.
\fancyname{} introduces the concepts of primary cycle compression (PCC) and learning residuals derived from simple smoothing average techniques. The former addresses the issue of scalar breakdown of dot-product attention and bypasses the high computational complexity of Transformer-based architectures, while the latter helps to incorporate prior knowledge and thus improves prediction accuracy. \fancyname{} allows models to naturally adapt to concept drift and provides robust and explainable fallback behavior to statistical methods, both of which are highly desirable characteristics for real-world application in critical infrastructure systems. 

\fancyname{} can substantially improve current state-of-the-art Transformer-based models for time series forecasting.
We perform extensive evaluation of our approach on three representative model architectures on a variety of of datasets. 
While our experiments focus mainly on Transformer-based architectures, \fancyname{} in principle be easily incorporated as add-on into any kind of deep-learning based models.

In our results, we cannot confirm the superiority of neither one of the current SOTA Transformer-based architectures, nor of MLP-based approaches such as NHiTS, that were designed to overcome the drawbacks in computational demand of Transformers.
We hypothesize that published results showing a clear advantage of any of these approaches originate from extensive hyperparameter tuning specific to the respective datasets. Contrary to that, when used in practice, the choice of best model varies. Our results demonstrate that regardless of the choice of model architecture, \fancyname{} can significantly improve forecasting accuracy. But more importantly, it drastically reduces demand in compute resources with respect to training time and energy consumption, thus providing a viable and useful approach to bringing state-of-the-art time series forecasting from theoretical studies into practical application.

\bibliographystyle{ieeetr}
\bibliography{LoadForecasting}

\begin{thebibliography}{10}

\bibitem{lara2021experimental}
P.~Lara-Benitez, M.~Carranza-Garcia, and J.~C. Riquelme, ``An experimental
  review on deep learning architectures for time series forecasting,'' {\em
  International Journal of Neural Systems}, vol.~31, no.~03, p.~2130001, 2021.

\bibitem{vaswani_attention_2017}
A.~Vaswani, N.~Shazeer, N.~Parmar, J.~Uszkoreit, L.~Jones, A.~N. Gomez,
  L.~Kaiser, and I.~Polosukhin, ``Attention {Is} {All} {You} {Need},'' {\em
  arXiv:1706.03762 [cs]}, Dec. 2017.
\newblock arXiv: 1706.03762.

\bibitem{zhou_informer_2021}
H.~Zhou, S.~Zhang, J.~Peng, S.~Zhang, J.~Li, H.~Xiong, and W.~Zhang,
  ``Informer: Beyond efficient transformer for long sequence time-series
  forecasting,''

\bibitem{wu_autoformer_nodate}
H.~Wu, J.~Xu, J.~Wang, and M.~Long, ``Autoformer: Decomposition transformers
  with auto-correlation for long-term series forecasting,'' {\em Advances in
  Neural Information Processing Systems}, vol.~34, pp.~22419--22430, 2021.

\bibitem{zhou_fedformer_2022}
T.~Zhou, Z.~Ma, Q.~Wen, X.~Wang, L.~Sun, and R.~Jin, ``{FEDformer}: Frequency
  enhanced decomposed transformer for long-term series forecasting,''
  no.~{arXiv}:2201.12740.
\newblock version: 3.

\bibitem{schwartz2020green}
R.~Schwartz, J.~Dodge, N.~A. Smith, and O.~Etzioni, ``Green ai,'' {\em
  Communications of the ACM}, vol.~63, no.~12, pp.~54--63, 2020.

\bibitem{wen2022transformers}
Q.~Wen, T.~Zhou, C.~Zhang, W.~Chen, Z.~Ma, J.~Yan, and L.~Sun, ``Transformers
  in time series: A survey,'' {\em arXiv preprint arXiv:2202.07125}, 2022.

\bibitem{li2019enhancing}
S.~Li, X.~Jin, Y.~Xuan, X.~Zhou, W.~Chen, Y.-X. Wang, and X.~Yan, ``Enhancing
  the locality and breaking the memory bottleneck of transformer on time series
  forecasting,'' {\em Advances in neural information processing systems},
  vol.~32, 2019.

\bibitem{liu2021pyraformer}
S.~Liu, H.~Yu, C.~Liao, J.~Li, W.~Lin, A.~X. Liu, and S.~Dustdar, ``Pyraformer:
  Low-complexity pyramidal attention for long-range time series modeling and
  forecasting,'' in {\em International Conference on Learning Representations},
  2021.

\bibitem{zhang2022crossformer}
Y.~Zhang and J.~Yan, ``Crossformer: Transformer utilizing cross-dimension
  dependency for multivariate time series forecasting,'' in {\em The Eleventh
  International Conference on Learning Representations}, 2022.

\bibitem{woo2022etsformer}
G.~Woo, C.~Liu, D.~Sahoo, A.~Kumar, and S.~Hoi, ``Etsformer: Exponential
  smoothing transformers for time-series forecasting,'' {\em arXiv preprint
  arXiv:2202.01381}, 2022.

\bibitem{nie2022time}
Y.~Nie, N.~H. Nguyen, P.~Sinthong, and J.~Kalagnanam, ``A time series is worth
  64 words: Long-term forecasting with transformers,'' {\em arXiv preprint
  arXiv:2211.14730}, 2022.

\bibitem{zeng_are_2022}
A.~Zeng, M.~Chen, L.~Zhang, and Q.~Xu, ``Are {Transformers} {Effective} for
  {Time} {Series} {Forecasting}?,'' Aug. 2022.
\newblock arXiv:2205.13504 [cs].

\bibitem{oreshkin2019n}
B.~N. Oreshkin, D.~Carpov, N.~Chapados, and Y.~Bengio, ``N-beats: Neural basis
  expansion analysis for interpretable time series forecasting,'' {\em arXiv
  preprint arXiv:1905.10437}, 2019.

\bibitem{challu2023nhits}
C.~Challu, K.~G. Olivares, B.~N. Oreshkin, F.~G. Ramirez, M.~M. Canseco, and
  A.~Dubrawski, ``Nhits: Neural hierarchical interpolation for time series
  forecasting,'' in {\em Proceedings of the AAAI Conference on Artificial
  Intelligence}, vol.~37, pp.~6989--6997, 2023.

\bibitem{taubert2023propulate}
O.~Taubert, M.~Weiel, D.~Coquelin, A.~Farshian, C.~Debus, A.~Schug, A.~Streit,
  and M.~Götz, ``Massively parallel genetic optimization through asynchronous
  propagation of populations,'' 2023.

\bibitem{paszke2019pytorch}
A.~Paszke, S.~Gross, F.~Massa, A.~Lerer, J.~Bradbury, G.~Chanan, T.~Killeen,
  Z.~Lin, N.~Gimelshein, L.~Antiga, {\em et~al.}, ``Pytorch: An imperative
  style, high-performance deep learning library,'' {\em Advances in neural
  information processing systems}, vol.~32, 2019.

\end{thebibliography}

\end{document}